\documentclass[10pt,twocolumn,letterpaper]{article}

\usepackage{cvpr}
\usepackage{times}
\usepackage{epsfig}
\usepackage{graphicx}
\usepackage{amsmath}
\usepackage{amssymb}
\usepackage{multirow}
\usepackage{booktabs}
\usepackage{flushend}

\usepackage[breaklinks=true,bookmarks=false]{hyperref}

\cvprfinalcopy 

\def\cvprPaperID{****} 

\begin{document}

\title{Team RUC AI$\cdot$M$^3$ Technical Report at Activitynet 2020 Task 2:\\ Exploring Sequential Events Detection for Dense Video Captioning}
\author{Yuqing Song, Shizhe Chen, Yida Zhao, Qin Jin\thanks{Corresponding author.}\\
School of Information, Renmin University of China\\
{\tt\small \{syuqing, cszhe1, zyiday, qjin\}@ruc.edu.cn}
}

\maketitle

\begin{abstract}
Detecting meaningful events in an untrimmed video is essential for dense video captioning.
In this work, we propose a novel and simple model for event sequence generation and explore temporal relationships of the event sequence in the video.
The proposed model omits inefficient two-stage proposal generation and directly generates event boundaries conditioned on bi-directional temporal dependency in one pass.
Experimental results show that the proposed event sequence generation model can generate more accurate and diverse events within a small number of proposals.
For the event captioning, we follow our previous work \cite{chen_2019} to employ the intra-event captioning models into our pipeline system.
The overall system achieves state-of-the-art performance on the dense-captioning events in video task with 9.894 METEOR score on the challenge testing set.
\end{abstract}

\section{Introduction}
The task of dense video captioning \cite{ranjay_densecap} aims to describe videos with a sequence of sentences rather than a single caption as in traditional video captioning.
To generate informative video descriptions, it is important to first detect meaningful events in the untrimmed video.

\begin{figure}[t]
    \centering
    \includegraphics[width=\linewidth]{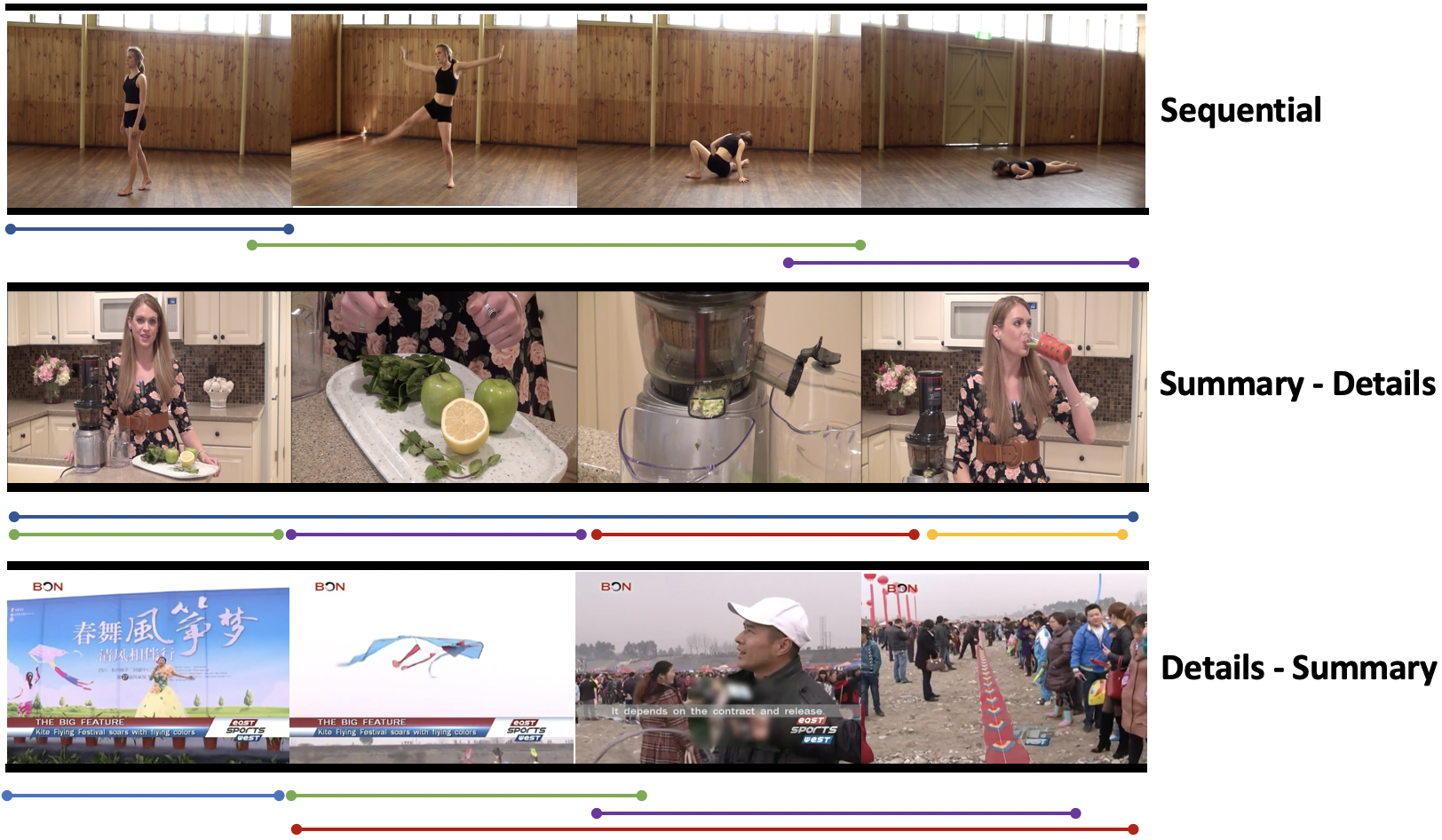}
    \caption{Examples of event sequences from ActivityNet dataset.}
    \label{fig:examples}
\end{figure}

Previous works \cite{li_jointly,wang_bidirectional,mun_streamline} mainly adopt a two-stage method to detect the events, including the candidate proposal generation stage and proposal selection stage.
The sliding windows or neural networks such as SST \cite{buch_sst} are first used to propose a large amount of event candidates.
Then event classifiers are designed to predict event confidence for each candidate.
The event proposals with confidences higher than a fixed threshold will be selected as the final events.
There are two major drawbacks of such two-stage approach:
First, a large amount of candidates (about 1000) need to be generated to ensure covering all the possible events, which is not efficient and computationally expensive.
Second, it does not consider temporal relationships between the events which may lead to candidates with high redundancy.
However, the events sequence in the video usually follows temporal orders as shown in Figure~\ref{fig:examples}.
We make statistics on the events sequential orders in ActivityNet dataset \cite{ranjay_densecap} based on the start and end timestamp of each event.
As shown in Table~\ref{tab:statistics}, about 81.5\% of videos in AcvitityNet dataset contain events in a clearly sequential order (one after another), while 16.94\% of videos contain events in a ``Summary-Details'' order.
Only 1.16\% of videos contain events not in any order.
Therefore, the events detection can be viewed as a sequence generation problem to directly generate event boundaries one by one.

In this work, we propose a novel and simple event sequence generation model, which fully exploits bidirectional temporal dependency of each event to generate event boundaries directly.
It takes previous event as the input and predicts next event distribution over the whole video timeline at each decoding step conditioned on the encoded videos.
To exploit both the past and future event contexts, we generate the event sequence in both forward and backward directions and then fuse the distribution maps in bi-directions to generate final event boundaries.
Experiments on ActivityNet Captions dataset demonstrate the proposed event sequence generation model can generate more accurate and diverse events with much less redundancy.
With the generated events sequence, the intra-event captioning models with contexts as in our previous work \cite{chen_2019} are further employed to generate descriptions for each event.
To take advantages of different captioning models, we utilize a video-semantic matching model to evaluate and choose more relevant captions from different models for each event.
The whole dense video captioning pipeline achieves the state-of-the-art performance on the challenge testing set.

The paper is organized as follows. 
In Section~\ref{sec:model}, we describe the whole dense video captioning system, which contains the event sequence generation module, the event captioning module and the re-ranking module.
Section~\ref{sec:experiments} presents the experimental results and analysis. 
Finally, we conclude the paper in Section~\ref{sec:conclusion}.
\section{Dense Video Captioning System}
\label{sec:model}
The whole framework of our dense video captioning system in ActivityNet Challenge 2020 consists of four components: 1) segment feature extraction; 2) event sequence generation; 3) event caption generation; and 4) event and caption re-ranking.

\begin{table}
\centering
\caption{The statistics of temporal order for events sequences in ActivityNet dataset}
\vspace{4pt}
\label{tab:statistics}
\begin{tabular}{c|c|c|c} \toprule
Sequential & Sum.-Details & Details-Sum. & Other \\ \midrule
81.50\% & 16.94\% & 0.40\% & 1.16\% \\ \bottomrule
\end{tabular}
\end{table}

\begin{figure*}[t]
    \centering
    \includegraphics[scale=0.55]{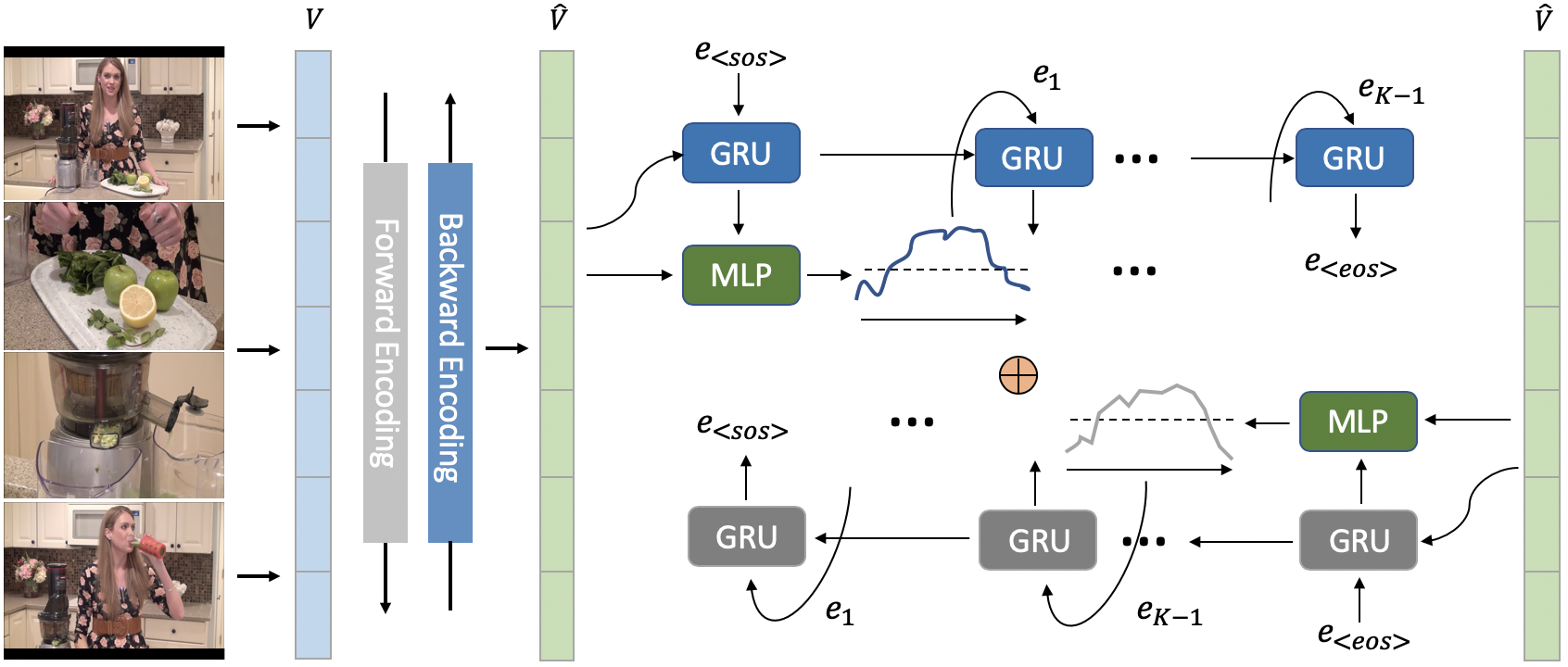}
    \vspace{4pt}
    \caption{Illustration of the proposed event sequence generation module.}
    \label{fig:model}
\end{figure*}

\subsection{Segment Feature Extraction}
\label{sec:segment-level}
Given an untrimmed video, we divide it into non-overlapping segments with 64 frames per segment.
Then we extract segment-level features as in our previous work \cite{chen_2019}, which includes: 1) Resnet200 \cite{he_resnet200} from image modality pretrained on ImageNet dataset;
2) I3D \cite{carreira_i3d} from motion modality pretrained on Kinetics dataset; and 3) VGGish \cite{hershey_vggish} from acoustic modality pretrained on Youtube8M dataset.
These three features are temporally aligned and are concatenated together as $v_t$ for the $t$-th segment. 
Therefore, the video is converted into $V=\{v_1,v_2,\cdots,v_T\}$, which is then used in the following modules.

\subsection{Event Sequence Generation}
To generate the event sequence with bidirectional video contexts, we further encode the segment-level features $V$ into context-aware features $\hat{V}$.
We employ a bidirectional GRU \cite{cho:gru} on the segment-level features sequence $V$ to capture the visual context in both forward and backward directions.
The hidden states in two directions at each encoding step $\overrightarrow{h_t}$ and $\overleftarrow{h_t}$ are concatenated and added to the segment-level feature $v_t$ followed by ReLU.
Therefore, the context-aware feature $\hat{v_t}$ for the $t$-th segment can be expressed as:
\begin{equation}
    \hat{v_t} = \mathrm{ReLU}([\overrightarrow{h_t}, \overleftarrow{h_t}] + v_t)
\end{equation}

Conditioned on the context-aware video features $\hat{V}$, we then generate events by another GRU one by one.
We represent the $i$-th event as a T-dimensional binary feature vector $e_i=\{x^i_0,x^i_1,\cdots,x^i_T\}$, where T is the number of video segments.
The value $x^i_t$ is set as 1 if the $t$-th segment is included in the $i$-th event interval, otherwise it is set as 0.
We utilize all-zeros vector as the special start event $e_{<sos>}$ and special end event $e_{<eos>}$.

We initialize the hidden state of the event decoder $\mathrm{GRU_e}$ with global video feature $\bar{\hat{v}}=\frac{1}{T}\sum_{t=0}^T \hat{v_t}$.
The event decoder uses previous generated event $e_{i-1}$ as input and predicts the $i$-th event distribution over the whole video timeline as follows:
\begin{eqnarray}
    h_i = \mathrm{GRU_e}(e_{i-1},h_{i-1}) \\
    p^i_t = \sigma(\mathrm{MLP}([h_i,\hat{v_t}])) \\
    e_i = \mathbb{I}(p^i_t >= 0.5)
\end{eqnarray}
where MLP is multilayer perceptron, $\mathbb{I}$ is the indicator vector and $\sigma$ is the sigmoid function.
$p^i_t$ denotes the probability of the $t$-th segment included in the $i$-th event.
The timestamp of $e_i$ is therefore represented as [$t_0$, $t_1$], where $p^i_{t_0}$ and $p^i_{t_1}$ are the first and last probability over 0.5.
In such way, the event decoder can generate the event sequence one by one until $e_i$ is the special end event $e_{<eos>}$.

The binary cross entropy is utilized to optimize the event distribution $p_i=\{p^i_0,p^i_1,\cdots,p^i_T\}$ as follows:
\begin{equation}
\label{eqn:binary_cross_entropy}
    \mathcal{L}_{ESG} = - \sum_{n=1}^N \sum_{i=1}^{K}\log P(e_i^n|e_{<i}^n,\hat{V}; \Theta)
\end{equation}
where $K$ is the number of events in the $n$-th video, and $\Theta$ represents all the learnable parameters in event sequence generation module.
For faster learning, we utilize the teacher forcing training strategy by feeding the ground truth event in each step.

In such forward generating direction, we generate the events sequence $<e_0,e_1,\cdots,e_K>$ one by one only depending on the past events, which ignores the future event contexts.
Therefore, we train another event generator with the whole video reversed, and generate the events sequence $<e_K,e_{K-1},\cdots,e_0>$ in the backward direction, which exploits future events for each event prediction.
Finally, we match corresponding events in forward and backward directions with the tiou over 0.5 and average the predicted event distributional vectors in two directions to acquire the final event boundaries.
Figure~\ref{fig:model} illustrates the framework of event sequence generation module.

\subsection{Event Caption Generation}
The context also plays an important role in event caption generation.
Besides the basic segment-level video features $V$ described in Section~\ref{sec:segment-level}, we also capture the contextual information in the whole video with RNN as in \cite{chen_2019}.
We train a LSTM on segment-level feature sequence $V$ and the objective function is to predict concepts for each segment.
We take the hidden state of the LSTM as context feature $v^c_t$ for the $t$-th segment, and it is further concatenated with $v_t$ for the following caption generation.

As analysed in our previous work \cite{chen_2019}, the intra-event captioning models are faster and perform better than the inter-event captioning models.
Therefore, similar to \cite{chen_2019}, we adopt intra-event captioning model with local contexts for the event caption generation.
We adopt a two-layer stacked GRU as the decoder.
The first GRU layer in the decoder is the attention GRU, which takes the previous generated caption word $y_{t-1}$ and previous hidden state in the second GRU layer $h_{t-1}^{2}$ as inputs to calculate a query vector $h_t^1$ as follows:
\begin{equation}
    h_t^1 = \mathrm{GRU_1}([y_{t-1};h_{t-1}^2],h_{t-1}^1)
\end{equation}
where $h_0^1$ is initialized as the mean pooling of video features in the current event concatenated with local contexts.
The query vector $h_t^1$ is utilized to select relevant temporal contexts with attention mechanism.
Then the second GRU layer predicts the next caption word with the temporal context $c_t$ as follows:
\begin{eqnarray}
    h_t^2 = \mathrm{GRU_2}([h_t^1;c_t],h_{t-1}^2) \\
    p(y_t|y_{<t}) = \mathrm{softmax}(W_d h_t^{2})
\end{eqnarray}
where $W_d$ is the word embedding matrix.

We firstly train the captioning model based on ground-truth events with cross entropy loss and then fine-tune the model with self-critical reinforcement learning algorithms \cite{rennie_selfcritic} with rewards from METEOR and CIDEr.

\subsection{Event and Caption Re-ranking}
In order to improve the system robustness and further improve performance, we train different captioning models and propose the following re-ranking approach to ensemble different models.

\vspace{4pt}
\textbf{Event Re-rank:}
Since the precision of event proposals is vital to the final dense captioning performance, we combine the events generated by our proposed ESG module with the proposals in our previous work \cite{chen_2019}.
We adopt the same proposal re-rank policy in \cite{chen_2019} to get the final events.

\vspace{4pt}
\textbf{Caption Re-rank:}
With the fixed event proposals, we further re-rank captions generated by different captioning models for each event.
To select more accurate and visual relevant captions, we train a video-semantic matching model \cite{faghri2017vse++} on the ActivityNet caption dataset to evaluate the qualities of generated captions.
Finally, we choose the best caption based on the predicted score and the number of unique words for each event.

\section{Experiments}
\label{sec:experiments}

\begin{table*}[ht]
\centering
\caption{Event detection performances including recall and precision at four thresholds of temporal intersection of unions (@tIoU) on the
ActivityNet Captions validation set.}
\vspace{4pt}
\label{tab:proposal_comparison}
\begin{tabular}{c|ccccc|ccccc}
\toprule
\multirow{2}{*}{Methods} & \multicolumn{5}{c}{Recall(@tIoU)} & \multicolumn{5}{|c}{Precision(@tIoU)} \\
\cmidrule{2-11}
 & 0.3 & 0.5 & 0.7 & 0.9 & Avg & 0.3 & 0.5 & 0.7 & 0.9 & Avg \\
\midrule
ESGN \cite{mun_streamline} & \textbf{93.41} & \textbf{76.4} & 42.4 & 10.1 & 55.58 & 96.71 & 77.73 & 44.84 & 10.99 & 57.57 \\
\midrule
Forward-direction & 92.90 & 76.32 & 42.07 & 10.74 & 55.51 & 97.08 & 78.59 & 43.73 & 10.96 & 57.59 \\
Backward-direction & 91.53 & 74.49 & 42.21 & 10.49 & 54.68 & 96.84 & 79.02 & \textbf{45.84} & 11.58 & 58.32 \\
\textbf{Bi-direction} & 92.35 & 75.51 & \textbf{43.49} & \textbf{11.81} & \textbf{55.79} & \textbf{97.17} & \textbf{79.33} & 45.68 & \textbf{12.40} & \textbf{58.65} \\
\bottomrule
\end{tabular}
\end{table*}

\begin{table*}[ht]
\centering
\caption{Dense video captioning results including Bleu@N (B@N), CIDEr and METEOR on ActivityNet Captions validation set.}
\vspace{4pt}
\label{tab:caption_comparison}
\begin{tabular}{l|cccccc|cccccc}
\toprule
\multirow{2}{*}{Methods} & \multicolumn{6}{c}{with GT proposals} & \multicolumn{6}{|c}{with generated proposals} \\
\cmidrule{2-13}
 & B@1 & B@2 & B@3 & B@4 & CIDEr & Meteor & B@1 & B@2 & B@3 & B@4 & CIDEr & Meteor \\
\midrule
SDVC \cite{mun_streamline} & 28.02 & 12.05 & 4.41 & 1.28 & 43.48 & 13.07 & 17.92 & 7.99 & 2.94 & 0.93 & 30.68 & 8.82 \\
\midrule
Cross-entropy & 24.76 & 13.05 & 6.43 & 3.22 & 54.10 & 12.42 & 16.65	& 9.40 & 5.35 & 3.05 & 20.91 & 9.39 \\
CIDEr.SC & 26.94 & 14.29 & 6.75 & 2.94 & 53.58 & 13.75 & 17.91 & 10.18 & 5.57 & 2.88 & 20.44 & 10.38 \\
METEOR.SC & 26.16 & 13.80 & 6.49 & 2.85 & 47.29 & 14.53 & 15.60 & 8.82 & 4.87 & 2.59 & 13.28 & 10.70 \\
Re-ranking & 25.84 & 13.77 & 6.60 & 2.93 & 42.86 & 15.00 & 16.59 & 9.65 & 5.32 & 2.91	& 14.03	& 11.28 \\
\bottomrule
\end{tabular}
\end{table*}

\begin{table}
\centering
\caption{The evaluation performance on the testing set of two submissions. The “official” denotes using official split for training and “enlarged” denotes enlarging training set.}
\vspace{4pt}
\label{tab:overall}
\begin{tabular}{l|c|c} \toprule
 & Official & Enlarged \\ \midrule
GT events on val & 15.00 & 16.10 \\
Generated events on val & 11.28 & 12.27 \\
Generated events on test & 9.336 & 9.894 \\ \bottomrule
\end{tabular}
\end{table}

\subsection{Dataset}
We utilize the ActivityNet Dense Caption dataset \cite{ranjay_densecap} for dense video captioning, which consists of 20k videos in total with 3.65 event proposals per video on average.
We follow the official split with 10,009 videos for training, 4,917 videos for validation and 5,044 videos for testing in the experiments except for our final testing submission.
In the final submission, we enlarge the training set with 80\% of validation set, which results in 14,009 videos for training and 917 videos for validation. 
The video in training set contains one set of event proposal segmentation while video in validation set contains two sets of proposal segmentation.

\subsection{Evaluation of Event Proposal Generation}
\textbf{Implementation Details:}
We set the hidden units of GRU as 512.
There are two hidden layers in MLP with ReLU activation.
The maximum length of event sequence is set as 8.
Dropout of 0.5 is adopted to avoid the over-fitting.
We train the event proposal generation module for 30 epochs with the mini-batch size 8 videos and the learning rate 1e-4.

\vspace{4pt}
\textbf{Evaluation Metrics:}
We evaluate the performance of predicted events by measuring the recall and precision of proposals which have tiou 0.3, 0.5, 0.7 and 0.9 with the ground-truth.

\vspace{4pt}
\textbf{Experimental Results:}
Table~\ref{tab:proposal_comparison} presents our performance for event proposal generation.
Our proposed bi-directional event sequence generation model performs better than the two-stage method \cite{mun_streamline}.
Furthermore, it is simpler and more efficient.
It generates 2.89 events per video on average, and the self-tiou of them is 0.07, which is close to the ground-truth events with self-tiou 0.05.
Generating the event sequence in forward and backward directions achieve competitive proposal performances.
Combining the two directions achieves the best performance on both average recall and precision, which shows the past and future events are both helpful for the current event prediction.

\subsection{Evaluation of Event Captioning}
\textbf{Implementation Details:}
For the caption decoder, we set the hidden units of GRU as 1024.
The dimensionality of word embedding layer is set to 300.
We initialize the embedding matrix with the pre-trained Glove \cite{pennington:glove}.
We train the captioning model for 30 epochs and select the model with best METEOR score on the validation set.

For the video-semantic matching model, the dimensionality of video-sentence joint embedding space is set as 1024.
Contrastive ranking loss with hard negative mining \cite{faghri2017vse++} is utilized for training.

\vspace{4pt}
\textbf{Evaluation Metrics:}
To purely evaluate the event captioning performance, we fix event proposals as the ground-truth proposals.
We employ the official evaluation process \cite{ranjay_densecap} with tiou threshold of 0.9 since we utilize the ground-truth proposals, and evaluate on common captioning metrics including BLEU, METEOR and CIDEr.
When evaluating the caption performance of generated events, we compute the caption performance for proposals possessing tiou 0.3, 0.5, 0.7 and 0.9 with the ground-truth.

\vspace{4pt}
\textbf{Experimental Results:}
Table~\ref{tab:caption_comparison} shows our dense captioning performances on the ground-truth events and generated events.
The intra-event captioning model trained with cross-entropy loss has achieved competitive performance with the METEOR 12.42.
Fine-tuning the model with the rewards computed by CIDEr and METEOR metrics in reinforcement learning framework further improves the captioning model significantly.
Ensembling various captioning models with caption re-ranking achieves additional improvements over all the single-models.
Compared with the performance on ground-truth events, the captioning performances on the generated proposals are much inferior, which infers the importance of event proposals generation.

The performances of our last two submitted models are presented in Table~\ref{tab:overall}.
In our final submission, we enlarge the training data with 80\% of validation set.
More training data brings substantial improvement, and our model achieves 9.894 METEOR score on the testing set.
\section{Conclusion}
\label{sec:conclusion}
In this work, we explore the temporal order of the event sequence in the video.
With fully exploiting temporal dependence in two directions, we propose a novel and simple event sequence generation model without traditional two-stage process.
For event captioning, we adopt the intra-event captioning models as our previous work \cite{chen_2019} and employ a video-semantic matching model to re-rank captions for each event.
Our proposed system achieves the state-of-the-art performance on the dense video captioning challenge 2020. 
In the future, we will further explore the coherence of multiple captions for the event sequence.

{\small
\bibliographystyle{ieee_fullname}
\bibliography{reference}
}

\end{document}